\documentclass[conference]{IEEEtran}
\IEEEoverridecommandlockouts
\usepackage{amsmath,amssymb,amsfonts}
\usepackage{algorithmic}
\usepackage{graphicx}
\usepackage{subfigure}
\usepackage{doi}
\usepackage{todonotes}
\usepackage{textcomp}
\usepackage{xcolor}
\usepackage[backend=biber]{biblatex}
\usepackage{bbm}
\def\BibTeX{{\rm B\kern-.05em{\sc i\kern-.025em b}\kern-.08em
    T\kern-.1667em\lower.7ex\hbox{E}\kern-.125emX}}
\makeatletter
\newcommand{\linebreakand}{\end{@IEEEauthorhalign}
\hfill\mbox{}\par
\mbox{}\hfill\begin{@IEEEauthorhalign}
}
\makeatother

\begin{document}

\title{On Hardening DNNs against Noisy Computations
}
\author{\IEEEauthorblockN{Xiao Wang, Hendrik Borras, Bernhard Klein, Holger Fröning}
	\IEEEauthorblockA{\textit{HAWAII Lab, Heidelberg University, Germany} \\
		\{xiao.wang, hendrik.borras, bernhard.klein, holger.froening\}@ziti.uni-heidelberg.de}
}

\maketitle
\begin{abstract}
The success of deep learning has sparked significant interest in designing computer hardware optimized for the high computational demands of neural network inference.
As further miniaturization of digital CMOS processors becomes increasingly challenging, alternative computing paradigms, such as analog computing, are gaining consideration.
Particularly for compute-intensive tasks such as matrix multiplication, analog computing presents a promising alternative due to its potential for significantly higher energy efficiency compared to conventional digital technology. 
However, analog computations are inherently noisy, which makes it challenging to maintain high accuracy on deep neural networks.
This work investigates the effectiveness of training neural networks with quantization to increase the robustness against noise.
Experimental results across various network architectures show that quantization-aware training with constant scaling factors enhances robustness. 
We compare these methods with noisy training, which incorporates a noise injection during training that mimics the noise encountered during inference.
While both two methods increase tolerance against noise, noisy training emerges as the superior approach for achieving robust neural network performance, especially in complex neural architectures.
\end{abstract}

\begin{IEEEkeywords}
Analog noise, Robustness, Quantization, Noisy training.
\end{IEEEkeywords}

\section{Introduction}\label{sec:introduction}

In the past decade, deep neural networks (DNNs) have been widely employed to address a variety of real-world problems, from fields such as image, signal and speech processing, natural language processing, and autonomous systems. 
Their ability to achieve remarkable performance on challenging tasks has made them a popular choice in various applications. 
DNNs are parametric models characterized by a vast number of weights, necessitating powerful specialized processors, such as GPUs, for training and evaluation. 
However, the rapid growth in the size of DNNs introduces new challenges related to memory-footprint, computation, and power consumption, particularly for deployment in resource-constrained embedded or mobile devices~\cite{JMLR:v25:18-567,compression_noisy}. 
This growing complexity contradicts the practical application of DNNs to real-world problems, as inference with these models often leads to unacceptable latency or excessive energy consumption in battery-powered devices.

In this regard, multiple recent studies have turned to alternative computing technologies, such as analog computing, for instance based on analog electrical computations~\cite{mixed-signal-computing,Kuhn2023}, analog optical computations~\cite{DNN-nanophotonic-circuits} and similarly emerging memory technologies such as resistive RAM~\cite{resistive-RAM, Emonds2024}.
In these applications, the computations inside  DNNs are executed in the analog domain with weights being represented by analog quantities, e.g. electrical voltage~\cite{preciseDNN-on-analog}, photons~\cite{DNN-nanophotonic-circuits, brckerhoffplckelmann2024probabilistic}, or conductance~\cite{NN-inference-phase-memory}.
Analog hardware has the potential to offer substantial improvements in energy efficiency, potentially exceeding two orders of magnitude compared to digital hardware~\cite{NN-inference-phase-memory,noisy_machine}.
However, analog quantities are subject to the noise inherently present in their physical components.
Depending on the technology, the weights may be written with some error~\cite{denoise-bayesian}.
Each weight readout can be overlaid with some noise, or the arithmetic operations themselves may also be fraught with noise.
In other words, analog computations are inherently noisy, and without countermeasures, this noise can negatively affect prediction accuracy.

To use such noisy hardware, methods to improve robustness against unreliable computations are essential.  From an algorithmic perspective, previous empirical work has shown that noise injection during training can lead to improvements in the noise resilience of analog computing devices~\cite{noisy_machine, noisy_bayes, NN-inference-phase-memory, analog-mixed-signal}. For instance, Gaussian noise is widely used in the training process of DNNs to improve the robustness~\cite{NN-inference-phase-memory, noisy_machine,analog-mixed-signal} during inference. Noisy Machines~\cite{noisy_machine} explores the use of knowledge distillation from a digitally trained teacher network to a student network with noise injection, while BayesFT~\cite{noisy_bayes} proposes different types injecting noise, including Bernoulli noise, Gaussian noise, and Laplace noise, to enhance the robustness of analog neural networks. Similarly, training as a counter measure to improve robustness against noise while slowly exposing a neural network architecture to an increasing amount of noise has proven to be effective~\cite{bernhard-incremental}. Moreover, \citeauthor{compression_noisy}~\cite{compression_noisy} have explored neural network compression techniques such as pruning and knowledge distillation on noisy storage devices.

Quantization is widely recognized as an effective method for compressing models on deterministic hardware; moreover, as analog hardware also faces practical limitations regarding precision, and quantization is implicitly required.
Quantization is also a good way to simulate the noise in analog-to-digital (ADC) and digital-to-analog (DAC) converters for neural network accelerators. 
By mapping weights and/or activations from floating-point representations into "bins" (lower precision formats, such as 8-bit or 4-bit integers), quantized neural networks may stay sturdy when models are perturbed by noise.
Similarly, is has been shown previously that that quantization can either improve or degrade adversarial robustness depending on the attack strength~\cite{quant_adv}.
In general, techniques such as quantization-aware training (QAT) have demonstrated significant success in mitigating quantization errors and enhancing generalization behavior in models~\cite{compress-quant-distill}. 
Given this promising body of previous work, it would be valuable to evaluate to which extend QAT improves the robustness against noise inherently present in analog computations, particularly as best to our knowledge, this question has not been explored in the literature before.

In the following, we thus explore the robustness of quantized neural networks in the presence of noisy computations.
As a backdrop, we also investigate models trained with noise injection to provide a quantitative comparison.

To quantify and compare the robustness as a key performance metric, we employ the \textit{midpoint noise level $\mu$} as proposed in~\cite{midpoint}.
Our contributions can be summarized as follows: 
\begin{enumerate}
	\item We investigate the \emph{robustness by quantization} against analog noise and compare robustness across different precision formats as well as constant and dynamic scaling. 
	\item Similarly, we investigate noisy training methods, which simulates the conditions of a real noisy analog device, compelling the network to adapt to noise during the training process.
	\item Finally, we present an overall evaluation, comparing quantization methods and noisy training techniques across different architectures.
\end{enumerate}

Empirically, we evaluate the effectiveness of our methods on image classification tasks using models trained on CIFAR-10. Our evaluation shows that quantization and noisy training can all enhance the robustness to different extents.

 \section{Related Work}\label{sec:related_work}
Noisy Machines~\cite{noisy_machine} models generic non-volatile memory (NVM) cell noise as an additive zero-mean independent and identically distributed (\textit{i.i.d.}) Gaussian noise term on the weights $\boldsymbol w_i$ of the model in each particular layer $l$: $\Delta \boldsymbol w_i \sim \mathcal{N} (\Delta \boldsymbol w_i; 0, \sigma^2_{N,l} \mathbf{I})$, where $\mathbf{I}$ is the identity matrix and $\sigma_{N,l} $ is the noise standard deviation of layer $l$.
Moreover, they examine the training with injected Gaussian noise to increase robustness against such analog computations and proposed knowledge distillation as a further extension to increase robustness. BayesFT~\cite{noisy_bayes} adopts a memristor perturbation model, which considered multiple factors resulting in the memristance drifting. Specifically, the drifting term is applied to each neural network parameter $\boldsymbol w ^{\prime}_i \leftarrow \boldsymbol w_i e^\lambda, \lambda \sim \mathcal{N} (0, \sigma^2)$, where $\boldsymbol w^{\prime}_i$ is the drifted neural network parameter.
Compared with Noisy Machines and BayesFT, injecting noise on weights, \textit{Walking Noise}~\cite{midpoint} focus on injection at the output activation, addressing combined noise from weight readout and the subsequent computations such as a dot product.

Previous work on quantization have primarily focused on its effects on DNNs' robustness to adversarial examples~\cite{quant_adv, defensive_quant,quant_adv_relative_robustness,robust_quant_one_rule_all,quant_bits_adv} and its impact on neural networks with different architectures~\cite{Resiliency_of_QNN} and quantization processes~\cite{robust_quant_one_rule_all}. 

\citeauthor{quant_bits_adv}~\cite{quant_bits_adv} indicate that robustness against adversarial attacks is non-monotonic in the number of bits. \citeauthor{quant_adv_relative_robustness} ~\cite{quant_adv_relative_robustness} find that quantization can improve a network’s resilience to adversarial attacks overall whilst causing negligible loss of precision. \citeauthor{quant_adv}~\cite{quant_adv} suggest that there is a critical adversarial attack strength at which quantization has little-to-no effect on accuracy. For attack strengths less than this critical strength, increasing weight precision improves accuracy by enabling more complex decision boundaries. At attack strengths greater than the critical strength, increasing precision causes a drop in accuracy stemming from decision boundaries being closer to data points. Defensive Quantization~\cite{defensive_quant} highlights that vanilla quantization suffers more from adversarial attacks due to the error amplification effect, where the quantization operation further enlarges the distance caused by amplified noise. They propose to control the Lipschitz constant of the network during quantization, such that the magnitude of the adversarial noise remains non-expansive during inference. 

Additionally, \citeauthor{Resiliency_of_QNN}~\cite{Resiliency_of_QNN} analyse the effects of quantization on feedforward deep neural networks and convolutional neural networks as their complexity varies.  This study also show that highly complex DNNs have the capability of absorbing the effects of severe weight quantization through retraining, but connection-limited networks are less resilient. To provide intrinsic robustness to the model against a broad range of quantization processes, Robust Quantization~\cite{robust_quant_one_rule_all} introduces a kurtosis regularization term, which is added to the model loss function. 
 \section{Methodology}
While analog accelerators promise to be orders of magnitude more energy efficient
than their digital counterparts, they are inevitably fraught with noise and non-linearities in their computations. 
In this section, we describe the metric utilized to evaluate the robustness of DNNs under noisy computations, as well as the methods employed as counter measures to losses in robustness due to noise.

\subsection{Robustness: midpoint noise level}
To quantify robustness of a DNN against injected noise, we measure the \textit{midpoint noise level $\mu$}; a metric developed in a previous work by \citeauthor{midpoint}~\cite{midpoint}.
It is defined as the injected noise at which the network achieves half of its maximum accuracy, precisely $\delta a = \frac{a_{max} - a_{min}}{2}$, with $a_{max}$ and $a_{min}$ being the maximally and minimally achieved accuracy, respectively. Under normal circumstances, $a_{min}$ equates to the random prediction accuracy of a given dataset. 
Equation \eqref{midpoint_fit} describes a scaled and shifted logistic function which is fitted to the observed data,
\begin{equation}\label{midpoint_fit}
	F(\sigma; \mu, s, \delta a, a_{min}) = \frac{2}{1+e^{(\sigma-\mu)/s}} \cdot \delta a + a_{min}
\end{equation}
with $\mu$ as specified before, $s$ the curve’s slope, and $\delta a$ and $a_{min}$ being the curve's scale and shift factor. Fitting a function to the data additionally gives us the ability to link data points with uncertainty information, which are a natural result of fluctuations during training. Thus the error estimate returned by the fit can be used to assess the reliability of the obtained result:
\begin{equation}\label{midpoint}
	\mu = \arg\min_{\mu, s, \delta a, a_{min}} \left\| \frac{F(\sigma; \mu, s, \delta a, a_{min}) -y(\sigma)}{\Delta y(\sigma)}\right\| ^2, \forall \sigma
\end{equation}
where $y(\sigma)$ and $\Delta y(\sigma)$ refer to the observed accuracy and uncertainty, respectively. 

This metric follows the intuition that a larger $\mu$ corresponds to higher robustness against noise. And is roughly equivalent to simply finding the data point closest to the threshold, where the investigated network achieves half of the maximum possible accuracy. However, the metric is more stable against fluctuations by using the entire data and incorporating statistical errors and additionally allowing to quantify how certain the resulting observation is. 

By injecting noise globally with the same intensity at all layers of the network, the \textit{midpoint noise level} $\mu$  can reflect the sensitivity of a network to noise. Furthermore, noise can also be injected exclusively at a single layer to probe how the internals of a network react to the noise. 

\subsection{Quantization}\label{quant_basic}
When applying quanitzation to DNNs, the goal is to reduce the precision of both the parameters $\boldsymbol w_i$, as well as the intermediate activation maps $\boldsymbol a_i$ to a lower precision, with minimal impact on the generalization power, in other terms accuracy of the model.  To do this, we first provide a quantization scheme that maps a floating point value to a quantized one, such as an integer, and then introduce how we adapt this scheme for training with backpropagation.

In this work we focus on the common quantization method, \textit{uniform quantization}, where the distance $s$ between quantization intervals is identical across all quantization levels. This distance $s$ is also called the \textit{scaling factor}.  Then, the quantization function can be defined as follows: 
\begin{equation}\label{eq:quant_uni}
	Q(x) =  \lfloor{\frac{x}{s}}\rceil + z,
\end{equation}
where  $Q$ is the quantization operator, $x$ is a real-valued input, $\lfloor{\cdot}\rceil$ is  the rounding operation (e.g. round to nearest or round to floor) and $z$ is the \textit{zero point}. The scaling factor and the zero point are used to map a floating point value to the integer grid, whose size depends on the bit-width $b$. The scaling factor can be defined manually as a hyperparameter, but it can also be defined according to the desired range of real values: 
\begin{equation}\label{eq:scale}
	s = \frac{\beta - \alpha}{2^b -1},
\end{equation}
where $[\alpha, \beta]$ denotes the bounded range that real values are clipped with.

\textbf{Symmetric and asymmetric quantization.} \textit{Symmetric quantization} is a simplified version of the general asymmetric case. Symmetric quantization restricts the zero point to 0, i.e. $-\alpha = \beta$. This reduces the computational overhead of dealing with the zero point offset. However,  \textit{asymmetric quantization} often results in a tighter clipping range and thus better representation of the actual value space, since the clipping range can be chosen to use the min/max of the input, i.e. $\alpha = x_{min}, \beta = x_{max}$, which results in the zero point non-zero.  This is especially important when the target weights or activations are imbalanced, e.g., the activations after a ReLU, which are always of positive value.

After the representation range $[\alpha, \beta]$ is determined, any values of $x$ that lie outside of this range will be clamped to its limit:
\begin{equation}\label{eq:clamp}
	q_l = \text{clamp}\Bigl(\lfloor{\frac{x}{s}}\rceil + z, q_{min}, q_{max}\Bigr),
\end{equation}
where $q_l$ is the quantization level , $[q_{min}, q_{max}]$ denotes the range of the integer grid, i.e. $\{0, \ldots, 2^b-1\}$ for \textit{unsigned} integers and $\{-2^{b-1}, \ldots, 2^{b-1}-1\}$  for \textit{signed} integers.
Here, $[q_{min}, q_{max}]$ are equivalent to the quantized values of $[\alpha, \beta]$.

Notably, the clipping procedure may incur a \textit{clipping error}~\cite{white_paper_quant_new}. 
To reduce the clipping error one can expand the quantization range by increasing the scaling factor. However, increasing the scaling factor leads to increased \textit{rounding errors}~\cite{white_paper_quant_new} as the rounding error is dependent on the range $\left[-\frac{1}{2}s, \frac{1}{2}s\right ]$. As such choosing the scaling factor $s$ becomes a trade-off.

\textbf{Static and dynamic quantization.}
Except for the methods for determining the clipping range $[\alpha, \beta]$, another important differentiator of quantization methods is \textit{when} the clipping range is determined. This range can be computed statically for weights, as the parameters are fixed during inference. However, the activation maps differ for each input sample. So, there are two approaches to quantizing activations: \textit{dynamic quantization}, and \textit{static quantization}. 

In dynamic quantization, this range is dynamically calculated for each activation map during runtime. This approach requires real-time computation of the signal statistics (min, max, percentile, etc.) which can have a significant overhead. However, dynamic quantization often results in higher accuracy as the signal range is exactly calculated for each activation.

Another quantization approach is static quantization, in which the clipping range is pre-calculated and static during inference. This approach does not add any computational overhead, but it typically results in lower accuracy compared to dynamic quantization. In this work, we mainly focus on static quantization. Specifically, we investigate \textit{constant scaling} and \textit{dynamic scaling}. In constant scaling, a single scaling factor is applied to all activations, whereas in dynamic scaling, each activation (or groups thereof) has its own scaling factor. We use the term \textit{dynamic scaling} to differentiate it from \textit{dynamic quantization}.

\textbf{Quantization-aware training.}
In this paper, we focus on quantization compatible with backpropagation, generally called quantization-aware training (QAT), which evaluates the impact of parameter quantization during training and adjusts parameters using training data to mitigate accuracy degradation, thus achieving higher accuracies than direct post-training quantization. A common method for implementing QAT is to insert fake quantization~\cite{quant_survey}, where values are stored in full precision but discretized during computation. Additionally, the Straight Through Estimator (STE)~\cite{STE} is used to address the non-differentiable quantization operator in backpropagation by approximating the rounding operation with an identity function.

\begin{figure}[hbtp]
	\centering
	\includegraphics[width=0.5\textwidth]{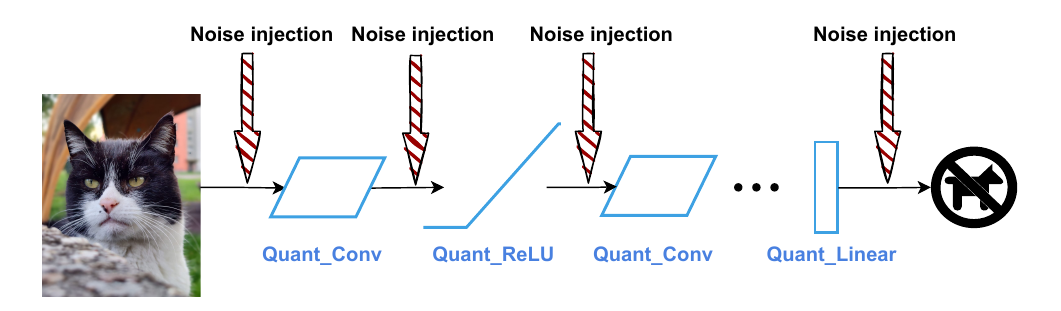}
	\caption{Global noise injection in a quantized neural network.}
	\label{fig:quant_noise_injection}
\end{figure}
A quantized convolutional neural network with global noise injection is depicted in Fig.~\ref{fig:quant_noise_injection}.

 \section{Experiments}
All experiments are conducted using the CIFAR-10 dataset~\cite{cifar-10}. We focus on three representative convolutional neural network architectures, LeNet-5~\cite{lenet5},  VGG~\cite{vgg} and ResNet~\cite{resnet}.
The simplicity of LeNet-5 allows us to isolate the effects of quantization without the added complexity of more modern, deeper networks. VGG and ResNet networks, in contrast, have significantly deeper architectures. This depth allows for the study of quantization effects across many layers, providing insights into how quantization impacts deeper networks. 

To quantize a neural network, we employ QAT, using the open-source framework \textit{Brevitas} \cite{brevitas} and follow the findings from~\citeauthor{quant_whitepaper_simulated}~\cite{quant_whitepaper_simulated}.
Specifically, we apply per-channel quantization on convolutional layers and per-tensor quantization on fully connected layers and activations. Uniform quantization is used for both weight tensors and activations.
For the activations, we explore both constant and dynamic scaling. The bit widths for weight tensors and activations are uniform.

As additive noise is often the primary type of noise in accelerators, we employ the quantified robustness metric \textit{midpoint noise level} $\mu$~\cite{midpoint} to asses the robustness of an architecture against noisy computation.
To inject noise at the activations  of a given neural network, a custom noise module is added after each layer. Similar to STE~\cite{STE}, the noise is only injected in the forward path, not the backward path. For all experiments, we inject Gaussian additive noise globally with the same intensity at all layers of the network.

\subsection{Robustness of QAT without noisy training}
\subsubsection{Results with LeNet-5}

\begin{figure*}[hbtp]
	\centering
	\subfigure[8-bit weights and activations]{\includegraphics[width=0.32\textwidth]{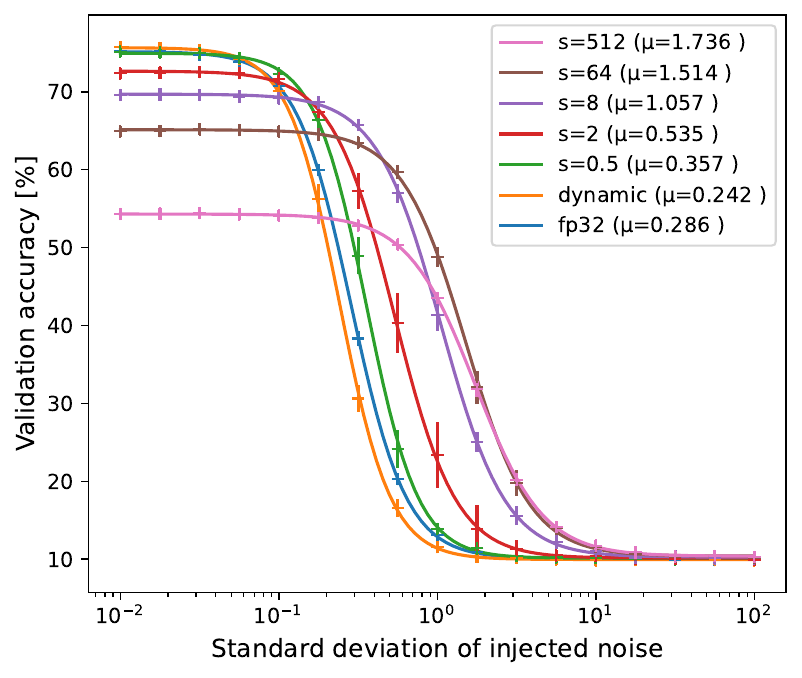}}
	\subfigure[4-bit weights and activations]{\includegraphics[width=0.32\textwidth]{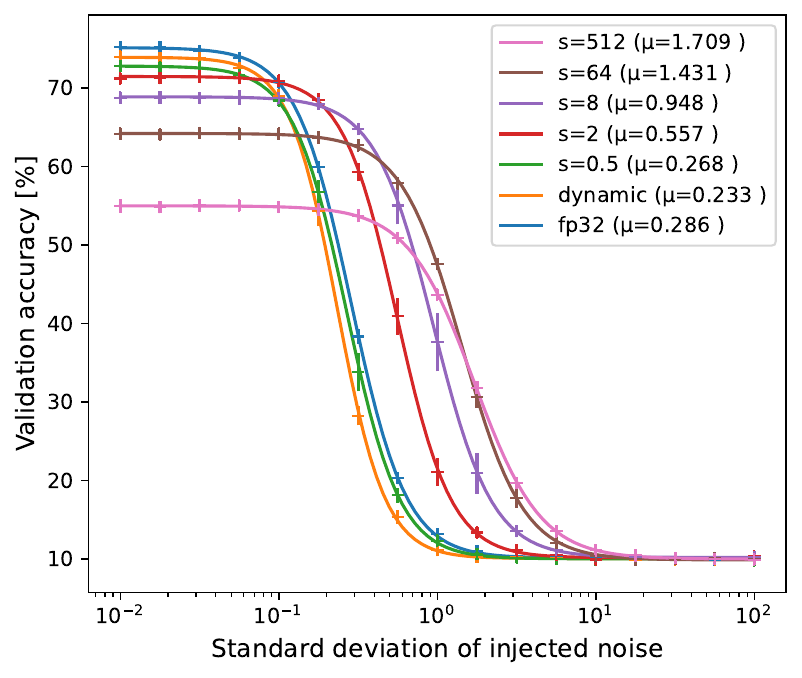}}
	\subfigure[Trade-off between accuracy and midpoint noise level $\mu$]{\includegraphics[width=0.32\textwidth]{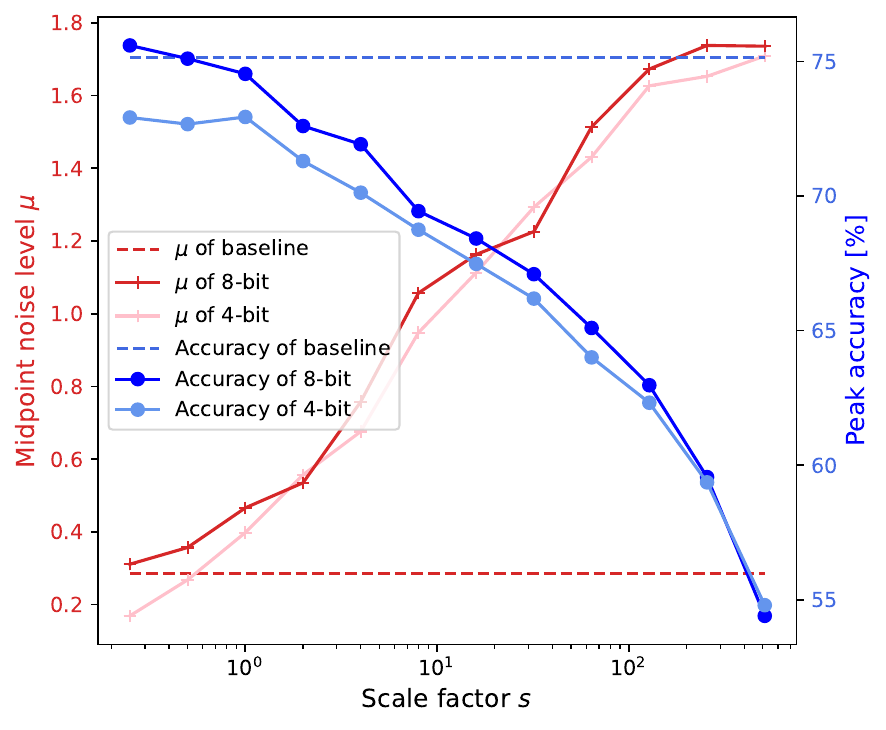}}
	\caption{Robustness of LeNet-5 on CIFAR-10, quantized with different bit widths, using either constant scaling (scaling factors $s$) or dynamic scaling on activations. Note that the x-axis is of logarithmic scale.}\label{fig:lenet-4and8bit}
\end{figure*}

In order to compare the effectiveness of quantization as a robustness method, all experiments are trained on the CIFAR-10 dataset without noise injection.
While the baseline is trained with floating-point precision.
Both the baseline and quantized models are trained using Adam~\cite{adam} with cosine learning rate decay~\cite{cosine}, and an initial learning rate of 0.001 (with batch size 128) for 500 epochs.
In order to isolate the effects of quantization, models are trained without batch normalization layers and regularization methods\footnote{While both batch normalization and regularization are important for generalization and thus maximizing test error, they are not used here to avoid interference with noise experiments. Future work will revisit noise and regularization in combination. A discussion on the impact of batch normalization on midpoint noise level can be found in~\cite{midpoint}.}. To observe the uncertainty of data points, each inference is repeated 10 times on different networks with different random weight initialization. The accuracies obtained for different bit widths and different scaling methods are shown in Fig.~\ref{fig:lenet-4and8bit}. As expected the model accuracy degrades with increasing noise for the baseline LeNet-5 and its quantized variants.

The clipping range is computed statically for weights, as the parameters are fixed during inference. For the activations, we compare dynamic and constant scaling.
As can be seen, the dynamic scaling (orange curves in Fig.~\ref{fig:lenet-4and8bit} (a) and (b)) can achieve almost the same peak accuracy as the non-quantized baseline model for both 4-bit and 8-bit. However, the curves lie left to the baseline curve, which means that the robustness of the dynamic scaling is slightly worse than the baseline. 

For constant scaling factors, we identify a trade-off between the peak accuracy
and noise robustness of a model for both 4-bit and 8-bit. Larger scaling factors result in larger midpoint levels and lower peak accuracies. In the low noise regime, dynamic scaling factors and small constant scaling factors can preserve the peak accuracies, but when the constant scaling factors are larger than 1, the accuracies drop significantly. In the high noise regime, larger constant scaling factors can achieve significantly better accuracies than dynamic scaling factors up to a scaling factor of 512. As introduced in Section \ref{quant_basic}, the rounding error then lies in the range of $[-\frac{1}{2}s, \frac{1}{2}s]$, thus increasing the scaling factor leads to increased rounding error and reduced top accuracy. However, with constant scaling, the rounding error can be potentially mitigated by the weights during training.

To better evaluate the trade-off between different quantization granularities, Fig.~\ref{fig:lenet-4and8bit}(c) shows the trade-off between the peak accuracy and midpoint noise level of 4-bit model and the 8-bit model. The blue solid curves show that the peak accuracy is partially linearly correlated with the scaling factor.
Experiments with models quantized to 16-bit did not yield better results than 8-bit. Surprisingly, the 8-bit model outperforms the 4-bit model on both accuracy performance and robustness performance. A possible explanation is that the 8-bit model can represent wider clipping ranges, resulting in smaller clipping errors and consequently improved fitting capabilities.
\begin{table*}[hbtp]
	\caption{Robustness of VGG-11 and ResNet-18 on CIFAR-10, quantized with  different bit widths and scaling factors. }
	\centering
	\begin{tabular}{c|cc|c|c} 
		\hline
		\textbf{Model} & \textbf{Bitwidths} & \textbf{Scaling factors} & \textbf{Peak accuracy (\%)} & \textbf{Midpoint noise level $\mu$ }\\
		\hline
		\rule[-1ex]{0pt}{2.5ex} &fp32 & - & \textbf{87.7} & 0.154 ($\pm$0.5\%) \\
		\cline{2-5}
		\rule[-1ex]{0pt}{2.5ex} &8-bit & dynamic & 87.2 & 0.024 ($\pm$0.1\%) \\
		\rule[-1ex]{0pt}{2.5ex} & & 0.5 & 84.3 & 0.145 ($\pm$0.2\%)\\ 
		\rule[-1ex]{0pt}{2.5ex} & & 1 & 82.2 & 0.2 ($\pm$0.2\%)\\ 
		\rule[-1ex]{0pt}{2.5ex} & & 2 & 76.8 & 0.222 ($\pm$0.3\%)\\ 
		\rule[-1ex]{0pt}{2.5ex} \textbf{VGG-11}& & 3 & 10.0 & 0.013 ($\pm$0.0\%) \\ 
		\cline{2-5}
		\rule[-1ex]{0pt}{2.5ex} &4-bit  & dynamic & 86.5 & 0.031 ($\pm$0.1\%) \\ 
		\rule[-1ex]{0pt}{2.5ex} & & 0.5 & 84.5 & 0.12 ($\pm$0.2\%)\\ 
		\rule[-1ex]{0pt}{2.5ex} & &  1 & 82.6 & 0.177 ($\pm$0.2\%)\\ 
		\rule[-1ex]{0pt}{2.5ex} & &  2 & 78.3 & \textbf{0.23 ($\pm$0.3\%)}\\ 
		\rule[-1ex]{0pt}{2.5ex} & &  3 & 10 & 0.010 ($\pm$0.1\%) \\
		\hline
		\hline
		\rule[-1ex]{0pt}{2.5ex} &fp32 & - & 87.0 & 0.495 ($\pm$0.2\%) \\
		\cline{2-5}
		\rule[-1ex]{0pt}{2.5ex} &8-bit & dynamic & \textbf{87.3} & 0.452 ($\pm$0.3\%) \\
		\rule[-1ex]{0pt}{2.5ex} & & 0.5 & 87.1 & 0.526 ($\pm$0.3\%)\\ 
		\rule[-1ex]{0pt}{2.5ex} & & 1 & 87.0 & 0. 557 ($\pm$0.3\%)\\ 
		\rule[-1ex]{0pt}{2.5ex} & & 4 & 86.5 & 0.577 ($\pm$0.3\%)\\ 
		\rule[-1ex]{0pt}{2.5ex} & & 8 & 85.7 & 0.625 ($\pm$0.4\%) \\ 
		\rule[-1ex]{0pt}{2.5ex} \textbf{ResNet-18}& &  10 & 10 & 0.05 ($\pm$2.5\%) \\
		\cline{2-5}
		\rule[-1ex]{0pt}{2.5ex} &4-bit  & dynamic & 86.8 & 0.281 ($\pm$0.3\%) \\ 
		\rule[-1ex]{0pt}{2.5ex} & & 0.5 & 86.5 & 0.492 ($\pm$0.4\%)\\ 
		\rule[-1ex]{0pt}{2.5ex} & &  1 & 86.7 & 0.605 ($\pm$0.3\%)\\ 
		\rule[-1ex]{0pt}{2.5ex} & &  4 & 86.5 & 0.657 ($\pm$0.5\%)\\ 
		\rule[-1ex]{0pt}{2.5ex} & &  8 & 86.5 & \textbf{0.665 ($\pm$0.3\%)} \\
		\rule[-1ex]{0pt}{2.5ex} & &  10 & 10 & 0.005 ($\pm$1.3\%) \\
		\hline
	\end{tabular}
	\label{table:vgg_quant}
\end{table*}

\subsubsection{Results with VGG and ResNet}
For experiments on VGG, we choose the achitecture VGG-11 using 50\% dropout on fully connected layers. We train both the full precision model (fp32) and quantized models for 500 epochs. The full precision model is trained with learning rate $\eta = 1\times10^{-3}$ , whereas the quantized models require a smaller learning rate of $\eta = 1\times10^{-4}$.   For experiments on ResNet, we choose the architecture ResNet-18, trained for 500 epochs with learning rate $\eta=0.01$ for both the full precision model and quantized models. The results are shown in Table \ref{table:vgg_quant}.

By comparing dynamic scaling with constant scaling, the advantages of constant scaling become evident. For both 8-bit and 4-bit of VGG-11 and ResNet-18, the midpoint noise levels with dynamic scaling are significantly smaller than those with constant scaling factors, up until the model completely collapses when the constant scaling factor reaches 3 for VGG-11 and 10 for ResNet-18, respectively.  When comparing models using dynamic scaling factors to the original floating-point model, it is clear that the dynamic scaling can nearly retain the original peak accuracy, while the midpoint noise levels $\mu$ are significantly lower.
Overall the conclusions here are identical to LeNet-5, previously.

\subsection{Experiments with Noisy Training.}
\begin{figure*}[hbtp]
	\centering
	\subfigure[LeNet-5, noisy training with static quantization]{\includegraphics[width=0.32\textwidth]{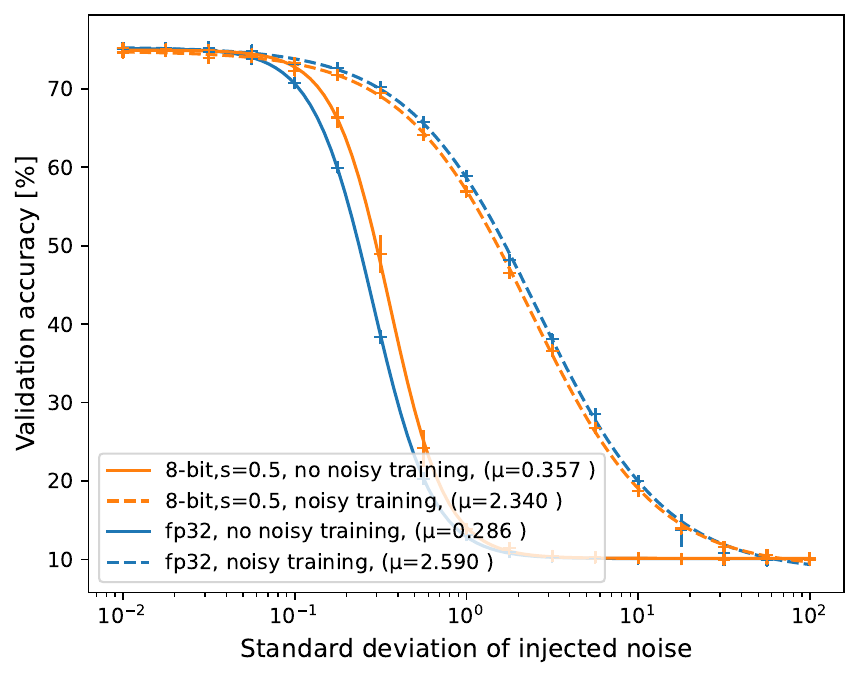}}
	\subfigure[Pareto analysis of LeNet-5 for different quantization levels.]{\includegraphics[width=0.32\textwidth]{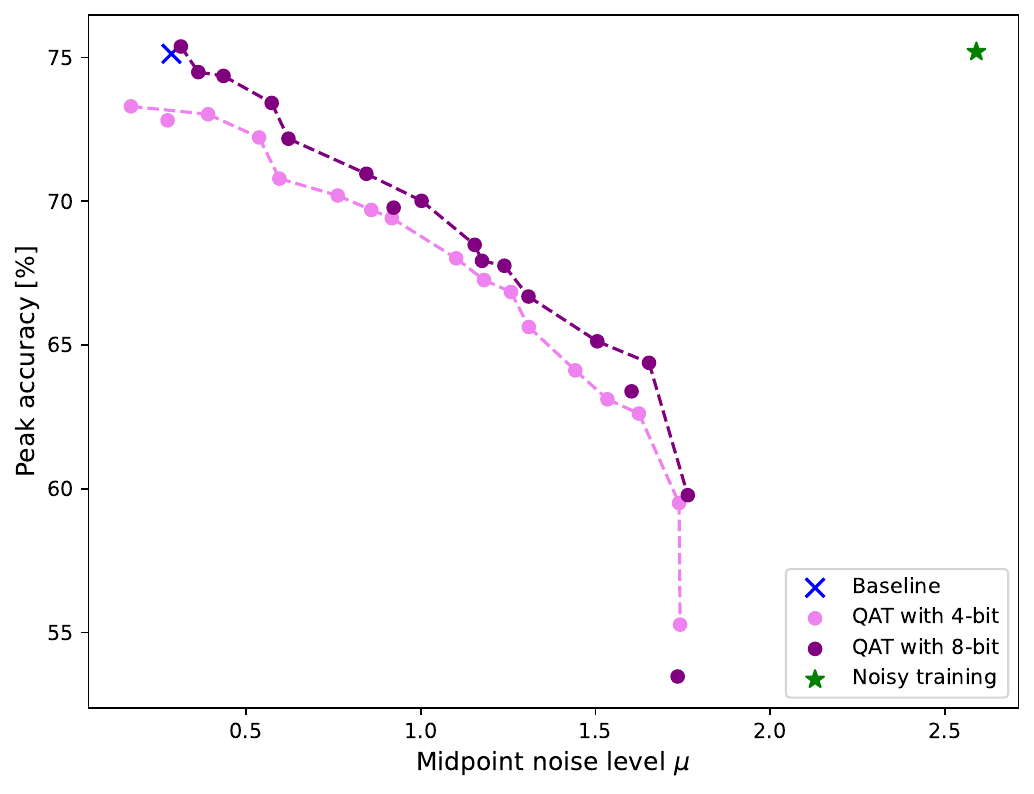}}
	\subfigure[ResNet-18, noisy training with static quantization.]{\includegraphics[width=0.32\textwidth]{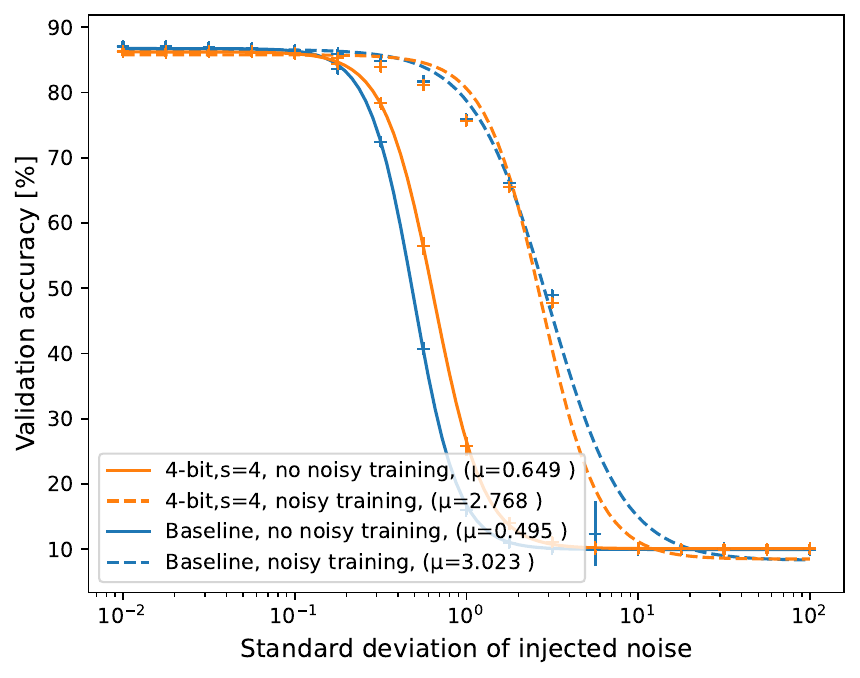}}
	\caption{Robustness of LeNet-5 and ResNet-18 on CIFAR-10 with noisy training, combined with quantization.}\label{fig:noisy}
\end{figure*}
Noise injection during training is a method used to expose the network to more realistic loss scenarios by randomly perturbing weights and activations. 
This simulates the conditions of a real noisy analog device, compelling the network to adapt to the noise during the training process. 
Ideally, training within the noisy analog systems leads to better empirical results, yet understanding the noise dynamics in analog hardware remains challenging.

To compare with noisy training techniques, we train the network while simultaneously injecting noise. 
The noise intensity is matched precisely to that encountered during inference (i.e., $\sigma_{\text{training}}= \sigma_{\text{inference}}$).  Results of LeNet-5 and ResNet-18 are shown in Fig.~\ref{fig:noisy}, (a) and (c).

From the results in Fig.~\ref{fig:noisy}, we observe that the dashed lines in each plot, representing noisy training combined with quantization, are closely matched, almost overlapping. This indicates that, regardless of whether the model is simple or complex, quantization methods can not significantly improve the robustness of models trained with noise injection.

The likely explanation for this inconsistency with the performance observed when models are trained without noise injection is that low precision quantization can degrade the fitting capability of models. This degradation helps explain why quantization does not improve the baseline performance of noisy training with floating-point 32-bit precision. Previous studies support these findings. For instance, \citeauthor{Resiliency_of_QNN}~\cite{Resiliency_of_QNN} discovered that highly complex DNNs can absorb the effects of severe weight quantization through retraining, while connection-limited networks exhibit less resilience. 

Although quantization methods do not improve the robustness of the baseline performance of noisy training, there is a positive aspect: quantized models trained with noise can achieve the same level of robustness as the floating point models while maintaining their accuracy. This finding suggests that it is feasible to deploy quantized models, which have smaller sizes, without sacrificing robustness.

\begin{table*}[hbtp]
	\caption{Overall comparison among architectures on CIFAR-10.}
	\centering
	\begin{tabular}{ c|c|c|c|c|cc} 
		\hline
		&  &  & &  & \multicolumn{2}{|c}{\textbf{Midpoint noise level $\mu$}} \\
		\textbf{Model}&\textbf{FLOPS (M)}&\textbf{Param (M)}&\textbf{Noisy layers}&\textbf{Peak accuracy (\%)}& \textbf{w/o Noisy Training}&\textbf{with Noisy Training}\\
		\hline
		\textbf{LeNet-5} & 0.66 & 0.06 & 12 & 75 & 0.286 &2.59\\
		\textbf{VGG-11} & 276.56 & 132.86 & 27 &87.7 & 0.154 &2.957\\
		\textbf{ResNet-18} & 37.53 & 11.69 &33 & 86.9 & 0.494 &3.023\\
		\hline
	\end{tabular}
	\label{table:flops}
\end{table*}
\subsection{Comparison among Architectures}
The experiments above reveal significant differences in the performance of various models. An overall comparison is provided in Table~\ref{table:flops}. In general, we make the following observations:

\begin{enumerate}
	\item \textbf{More complex models trade accuracy against robustness.} Without noise injection and quantization, VGG-11 and ResNet-18 achieve higher accuracies (87.7\% and  87.0\%, respectively) compared to LeNet-5 (75\%).  However, the midpoint noise level $\mu$ of VGG-11 (0.154) is significantly lower than LeNet-5 (0.286).  In general, complex models can achieve better accuracy, however, they also involve more multiplications and accumulations, overall leading to more noise involved and, consequently, less robustness. From Table~\ref{table:flops}, it is evident that VGG-11 has more noisy layers than LeNet-5.  However, with noisy training, complex models, which typically have a greater capacity to fit data, can learn to become more resilient to noise, thereby achieving higher robustness.
	
	\item \textbf{Skip connections reduce the amount of error accumulation, thereby improving robustness.} In contrast, ResNet-18 exhibits significantly more robustness, likely due to the used \emph{skip connections}. These connections allow the original stacked layers to focus on optimizing the residual mapping while the skip connections preserve the primary identity information, which is less exposed to noise.  Specifically, for one building block, three noisy layers are present in the original stacked layers, while only one or zero noisy layer is present in the skip connection, depending on whether downsampling is applied or not.  
	
	\item \textbf{Constant scaling is more robust than dynamic scaling.} For all three architectures, dynamic scaling is worse for robustness than constant scaling, and even worse than non-quantized \textit{fp32} values. To gain a more detailed understanding of this an in-depth analysis of activation distributions and effects related to training dynamics would be required, which is out-of-scope for this work.

\item \textbf{Larger scaling factors yield diminishing returns in terms of robustness with increasing model complexity.} On VGG-11 and ResNet-18, larger constant scaling factors result in only slight improvements in midpoint noise levels. For instance, for VGG-11 with 8-bit quantization and a constant scaling factor of 2, the midpoint noise level increases by 44.2\%, but accuracy drops from 87.7\% to 76.8\%, for ResNet-18 with 4-bit quantization and a constant scaling factor of 8, the midpoint noise level increases by 34.3\%.  In contrast, for LeNet-5 with 8-bit quantization and a constant scaling factor of 8, the midpoint noise level rises by over 200\%, while accuracy only decreases from 75\% to 70\%.  VGG-11 collapses when the scaling factor reaches 3, whereas LeNet-5 can function properly even with an extremely large scaling factors, such as 512.  Larger scaling factors come with larger quantization errors. The inferior robustness of VGG-11 can be attributed to the error amplification effect, where quantization errors are amplified through the layers of deep neural networks~\cite{defensive_quant}. 
\end{enumerate}

 \section{Conclusion}
Analog hardware holds the potential to significantly reduce the latency and energy consumption of neural network
inference, however, at the same time is imprecise and introduces noise within computations that limits accuracy in
practice.
In this work, we investigate the robustness of DNNs affected by analog noise during inference.
We employ quantization-aware training and noisy training techniques as robustness enhancing methods.

Our experimental results, conducted on the CIFAR-10 dataset with various models, including LeNet-5, VGG-11 and ResNet-18, indicate that quantization with constant scaling factors can significantly improve the robustness.
Large scaling factors, however, lead to accuracy loss.
Additionally, by comparing different architectures, we find that deeper network architectures are likely to suffer from error amplification, making them more sensitive to large scaling factors. 
Our results also show that when models are trained in an environment identical to the noisy conditions experienced during inference, robustness is significantly improved. 
Despite potential benefits of quantization, it does not improve the robustness of models when noise is injected during training. 
Overall, our findings highlight the importance of aligning training conditions with anticipated inference noise and suggest that targeted robustness strategies are essential to fully realize the benefits of analog hardware in practical, real-world applications.

\section*{Outlook}

This study underscores both the potential and the challenges of leveraging analog hardware to improve the speed and energy efficiency of neural network inference, despite the inherent computational noise it introduces.  
Our findings suggest several promising directions for future work to further enhance robustness in noisy environments.  
A key question is whether robust performance requires detailed knowledge of system noise or if approximate estimates of noise type and strength could suffice, simplifying practical implementations.  
Given the observed limitations of quantization alone, evaluating perturbation-based methods as complementary strategies may address weaknesses seen when training does not fully align with noisy inference conditions.  
Examining the combination of various robustness techniques could reveal potential synergies, strengthening noise resilience without significantly impacting performance.  

Additionally, further exploration of architectural factors such as network depth, width, residual connections, and attention mechanisms could clarify their influence on robustness, particularly since deeper architectures are prone to error amplification in noisy contexts.  
Improving robustness may also benefit from advanced sensitivity evaluation methods that go beyond techniques like walking noise, offering faster and more precise insights into the layer-specific impact of noise.  
Exploring robustness techniques adapted to each layer's unique sensitivity profile within a model may further enhance resilience.  
Advancing these research directions holds promise for developing robust, energy-efficient models optimized for noise-prone, analog hardware.

\printbibliography

\end{document}